\documentclass[letterpaper, 10 pt, conference]{ieeeconf}  

\IEEEoverridecommandlockouts                              

\usepackage{graphics} 
\usepackage{epsfig} 
\usepackage{mathptmx} 
\usepackage{times} 
\usepackage{amsmath} 
\usepackage{amssymb}  

\usepackage[backend=bibtex,style=ieee,natbib=true]{biblatex} 
\usepackage{accents}

\usepackage[switch]{lineno}

\usepackage{fancyhdr}
\fancypagestyle{withfooter}{
  
  \fancyfoot[C]{\footnotesize Accepted to the IEEE ICRA Workshop on Field Robotics 2024}
}

\usepackage{etoolbox}
\makeatletter
\patchcmd{\@makecaption}
  {\\}
  {.\ }
  {}
  {}
\makeatother

\title{\LARGE \bf
Real-World Deployment of a Hierarchical Uncertainty-Aware Collaborative Multiagent Planning System 
}

\author{Martina Stadler Kurtz$^{*1}$, Samuel Prentice$^{*1}$, Yasmin Veys$^{1}$, Long Quang$^{2}$, Carlos Nieto-Granda$^{2}$, \\Michael Novitzky$^{3}$, Ethan Stump$^{2}$, and Nicholas Roy$^{1}$
\thanks{*Equal contribution.}
\thanks{$^{1}$M.S. Kurtz, S. Prentice, Y. Veys, and N. Roy are with MIT CSAIL,
        {\tt\small \{mskurtz, prentice, vyeys, nickroy\}@mit.edu}.}%
\thanks{$^{2}$L. Quang, C. Nieto-Granda, and E. Stump are with the DEVCOM Army Research Laboratory (ARL), Adelphi, MD,
        {\tt\small \{long.p.quang.civ, carlos.p.nieto2.civ, ethan.a.stump2.civ\}@army.mil}.}%
\thanks{$^{3}$ M. Novitzky is with the United States Military Academy,
        {\tt\small michael.novitzky@westpoint.edu}.}%
\thanks{This research was sponsored by the Army Research Laboratory and was accomplished under Cooperative Agreement Number W911NF-17-2-0181. Their support is gratefully acknowledged.}
}

\begin{document}

\maketitle
\thispagestyle{withfooter}
\pagestyle{withfooter}

\begin{abstract}
We would like to enable a collaborative multiagent team to navigate at long length scales and under uncertainty in real-world environments. In practice, planning complexity scales with the number of agents in the team, with the length scale of the environment, and with environmental uncertainty. Enabling tractable planning requires developing abstract models that can represent complex, high-quality plans. However, such models often abstract away information needed to generate directly-executable plans for real-world agents in real-world environments, as planning in such detail, especially in the presence of real-world uncertainty, would be computationally intractable. In this paper, we describe the deployment of a planning system that used a hierarchy of planners to execute collaborative multiagent navigation tasks in real-world, unknown environments. By developing a planning system that was robust to failures at every level of the planning hierarchy, we enabled the team to complete collaborative navigation tasks, even in the presence of imperfect planning abstractions and real-world uncertainty. We deployed our approach on a Clearpath Husky-Jackal team navigating in a structured outdoor environment, and demonstrated that the system enabled the agents to successfully execute collaborative plans. 

\end{abstract}

\section{Introduction}

\begin{figure}
\centering
\includegraphics[width=0.9\linewidth]{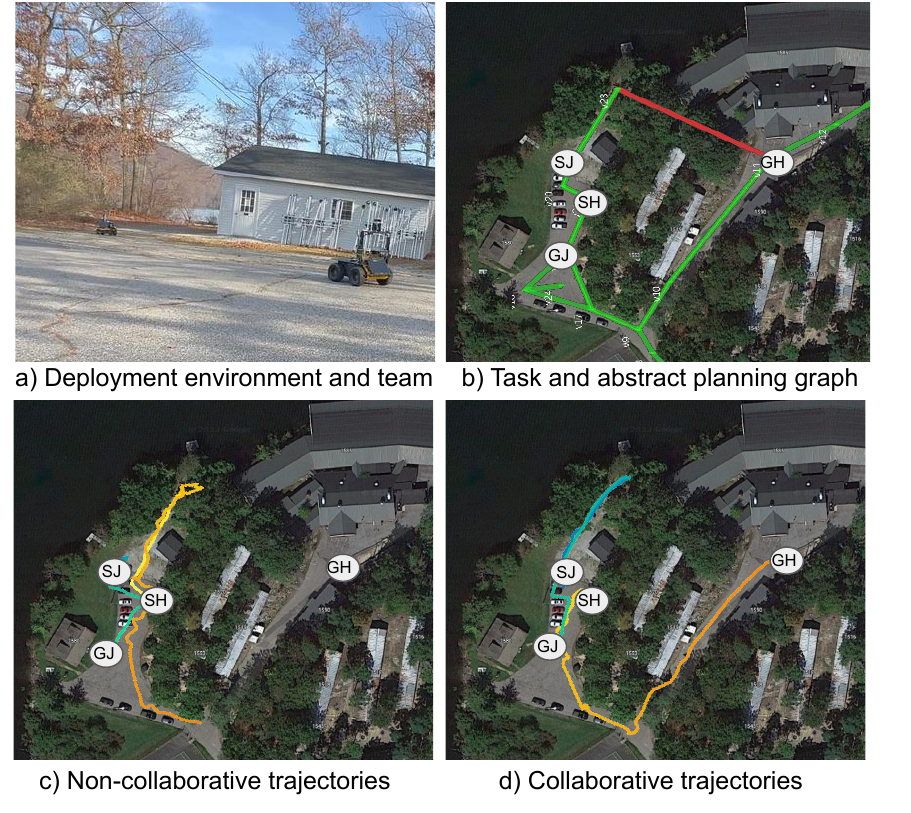}
\caption{ a) In this work, we discuss the deployment of a collaborative multiagent planner for a two agent team consisting of a Clearpath Jackal and a Clearpath Husky in a real-world structured outdoor environment. b) Consider the Jackal and Husky navigating from starts SJ and SH to goals GJ and GH, respectively, given a navigation graph (green), where one edge has unknown traversability (red). c) When the agents plan independently, the Husky attempts to navigate to the goal via the unknown edge, senses that the edge is untraversable, and then backtracks to the goal via a long, traversable path (trajectory shown in yellow-orange); meanwhile, the Jackal navigates directly to its goal (trajectory shown in blue-green).  d) When the agents plan to collaboratively minimize team makespan, the quicker Jackal diverts from the shortest path to its goal to sense the traversability of the unknown edge for the Husky, while the Husky waits in place for additional information. After sensing that the edge is untraversable, both the Jackal and the Husky navigate to their respective goals.}
\label{fig:front_fig}
\end{figure}

We would like to enable a team of robots to carry out collaborative missions autonomously in an unknown environment. For example, consider the team of agents navigating in the structured outdoor environment in Fig. \ref{fig:front_fig}-a, modeled as the graph in Fig. \ref{fig:front_fig}-b. When the agents plan collaboratively, they can take advantage of agent locations and traits to reduce the time to complete the whole mission as a team. In this example, the quicker Jackal robot senses the traversability of an unknown edge for the slower Husky robot. The observations from the Jackal enable the Husky to avoid slow, expensive backtracking when the edge is discovered to be untraversable, reducing the Husky plan cost and the overall team traversal time as compared to a non-collaborative baseline.

Unfortunately, generating good collaborative team plans can be computationally intractable as the size of the environment and team increases, since generating high-quality team plans under uncertainty requires answering three questions: \textit{what} is useful to explore given the team objective(s), by \textit{when} does the team need the explored information to improve planning performance, and \textit{who} should explore. Answering these questions directly requires a collaborative team to reason over a Partially Observable Markov Decision Process (POMDP) with large action and observation spaces, long time horizons, and delayed rewards, which is unlikely to be computationally tractable in real-world environments. To enable tractable, online planning for deployments in complex environments, we need to develop abstract models of the planning problem that reduce planning complexity.

In prior work, we developed a method for collaborative multiagent planning on abstract roadmaps, or motion graphs, where some graph edges were assumed to be probabilistically traversable \cite{ctp:stadler2023approximating}; the representation was selected to capture traversability uncertainty in unknown environments. First, we developed multiagent macro-actions that efficiently represented good collaborative team behaviors under uncertainty.
Second, we developed a macro-action pruning scheme that relied on an optimistic, Monte Carlo (MC) rollout-based value function approximator to estimate the effects of candidate future beliefs on team planning performance. Using the approach, we demonstrated the ability to find low-cost team policies that actively balanced between reducing task-relevant environmental uncertainty and efficiently navigating to goals. However, there is a trade-off to make between planning model complexity and real-world plan executability. As our models become more abstract to enable more efficient high-level planning, model properties that are needed to represent plan execution can become abstracted away. The challenges of executing abstract plans are exacerbated in real-world deployments, where real-world uncertainty, timing delays, and robot failures stress the relationship between the planning abstraction and real-world robot actions. For example, while the abstract motion graph model enabled efficient team planning, the output of the planner was graph-based macro-actions (i.e., sequences of motion, observation, and wait actions in the graph) that were not directly executable in a real-world environment.

To enable the real-world deployment of our collaborative multiagent planner, we developed a hierarchical planning system capable of generating abstract high-level plans and converting them into executable motor commands for real-world agents. The planner generated abstract, collaborative team plans, single-agent macro-action-based plans, and primitive action plans that could be directly executed on robots, all while maintaining plan consistency throughout the hierarchy and being robust to real-world uncertainty and disturbances.Our system was designed to handle both the theoretical and practical challenges of field robotics deployments. These challenges included grounding an abstract navigation graph and a graph-based team in a real-world environment, robustly executing abstract macro-actions in the presence of real-world disturbances, and coordinating teammates during the execution of variable-duration actions caused by both the planning abstraction and real-world uncertainty. We deployed our approach on a Clearpath Husky-Jackal team navigating in a structured outdoor environment, and showed that our system resulted in real-world collaborative team behaviors.

\section{Related Work}
The Canadian Traveller's Problem (CTP) has been used as a tool to study the complexity of navigation on graphs with stochastic edges \cite{papadimitriou1991shortest}. Various extensions to the original problem have been made, including to agents with remote sensing \cite{bnaya2009canadian}, and to graphs with specific structures that admit simple optimal policies \cite{ctp:routeplanning}. Other approaches use approximation strategies to generate high-quality policies for general CTP graphs \cite{ctp:high_quality}. Additionally, policies have been developed for multi-trial CTPs \cite{bnaya2015repeated} and multiagent CTPs \cite{bnaya2009canadian} for limited teams (i.e., teams where sensing agents have limited dynamics and no independent tasks). Most recently, we developed an approach for generating high-quality multiagent policies for CTPs which minimize team makespans for one ground vehicle/$n$ air vehicle teams \cite{ctp:stadler2023approximating}. 

Concurrently, other approaches have been developed for graph-based, risk-aware single-agent navigation under edge cost uncertainty. For example, Murphy and Newman \cite{murphy2012risky} developed an approach for generating stochastic graphs, given overhead imagery, and for generating plans in the graphs with bounded risk. Other approaches developed methods for online planning with uncertain edge costs and online edge cost sensing. Dey et. al. \cite{dey2014gauss} used a Monte Carlo based approach to generate agent policies given Gaussian process-based edge cost assignments, and Chung et. al. \cite{chung2019risk} developed a risk-aware search technique for planning with uncertain edge costs and sensing.

Finally, this work relies on core robot autonomy capabilities, including GPS-based state estimation, GPS-enabled Omnimapper-based mapping \cite{omnimapper}, EASL-based global planning \cite{easl}, and MPPI-based local planning \cite{mppi}. While our technique is not limited to the use of these specific techniques, the importance of a capable, reliable base autonomy system for field experiments cannot be overstated. 

\section{Approach}
In this section, we briefly discuss the collaborative multiagent planning approach developed in \cite{ctp:stadler2023approximating}. Then, we discuss the hierarchical planner developed for online, collaborative team planning in the real-world, and highlight challenges and opportunities from deploying the system.

\subsection{Collaborative Multiagent Planning Under Uncertainty}
In prior work, we formulated the problem of collaborative multiagent planning under uncertainty as a POMDP.  We modeled the environment as a stochastic navigation graph with unknown edge traversabilities. In the graph, nodes indicated locations, edges indicated possible paths, and each edge was assigned a traversability probability, or a probability that the edge was traversable during a given planning trial\footnote{This graph is similar to the stochastic graph developed in the Canadian Traveller's Problem (CTP), which was designed to model route finding in Canada; roads were assumed to be snowed in and impassable with some probability.}. The true traversability of each edge in the graph was assumed to be static throughout the planning trial, and could be observed by navigating to a node adjacent to the edge. Historically, the problem of planning on stochastic graphs was studied for single-agent navigation, and state-of-the-art techniques used Monte Carlo methods to generate high-quality agent policies by approximating expected future plan costs and trading off between exploring unknown, potentially low cost paths to the goal, and exploiting the current best known path to the goal \cite{eyerich2010high}. 

The goal of our prior work was to generate collaborative plans for teams navigating on stochastic graphs. Formally, the planning objective was to minimize the expected makespan for a team of agents navigating from individual starts to goals in the graph. In this problem, agents not only had to trade off exploring and exploiting paths to reach their own goals, but they also had to reason about whether or not they could sense information that would be helpful to a teammate. To increase the tractability of solving this complex multiagent POMDP, we designed collaborative macro-actions, or sequences of primitive actions, that were capable of representing good single-agent and collaborative plans in the unknown environment. The macro-actions were designed to capture intuitively useful agent behaviors in a single planning step; for example, we developed a macro-action that consisted of the actions an agent would take to navigate to and sense an unknown edge in the environment (see Fig. \ref{fig:macroaction}). By representing the collaborative sensing behavior as a single macro-action, we enabled the planner to reason about the benefit of sensing in a single planning step. Without macro-actions, the planner would have had to reason about the collaborative value of an agent navigating one step in the direction of the unknown edge, which is challenging because the benefit of the action is only obvious after a number of timesteps, when the agent reaches and senses the unknown edge. Along with the macro-action that modeled an agent navigating to sense the traversability of an edge, we also developed macro-actions that modeled an agent navigating via one or more known traversable edges to the goal, and an agent waiting in place for information to be sensed by a teammate.

\begin{figure}[t]
\centering
\includegraphics[width=0.8\linewidth]{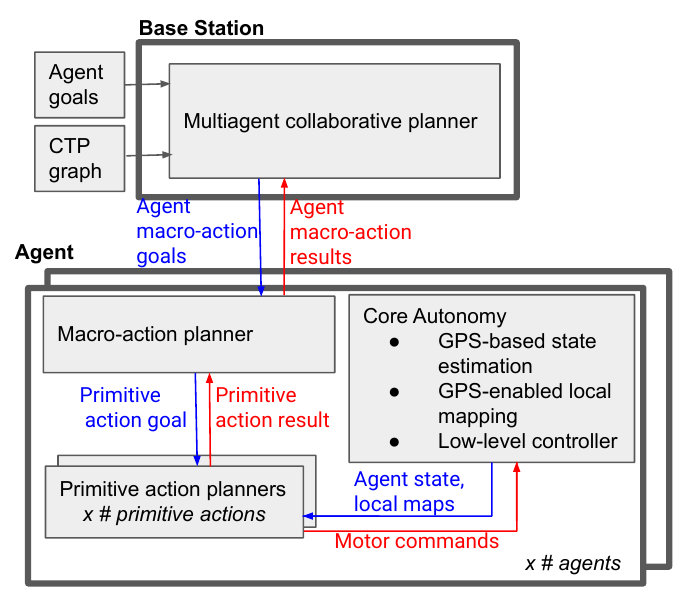}
\caption{Overview of of the hierarchical planning system. Multiagent planning, which occurred at a centralized base station, generated macro-actions for each agent in the team. The macro-actions were then sent to the individual agents, where they were processed and executed as sequences of primitive actions, using information from each agent's individual core autonomy pipeline.}
\label{fig:software_overview}
\end{figure}

While the macro-actions described reduced the depth of collaborative multiagent planning, there were still a large number of macro-actions to consider at every planning step. Most of the macro-actions, like those that sensed edges that were not on low cost paths to the goal, were unlikely to improve team planning performance. We developed macro-action-based value functions that enabled the planner to quickly prune macro-actions that were unlikely to lead to high-quality collaborative plans from the plan space. For additional details about the approach, please refer to \cite{ctp:stadler2023approximating}.

\subsection{Real-World Planning System Overview}
To deploy the system in the real-world, we developed a hierarchical planning system; an overview of the system is shown in Fig. \ref{fig:software_overview}. A stochastic navigation graph and agent goals were provided as inputs to the collaborative planner, which was run at a centralized base station. At each planning step, the centralized planner generated a macro-action for each agent to execute and communicated the macro-actions to the agents. Then, the agents executed the macro-actions onboard using a bi-level planner. At the higher planning level, the agent sequenced primitive action goals based on the macro-action, monitored primitive action outcomes, and had some limited ability to adjust primitive action goals based on primitive planning outcomes; for additional details, see Section \ref{sect:primitive_navigation_actions}. At the lower planning level, agents planned for and executed primitive actions, which included navigating along an edge, sensing the traversability of an edge, and waiting in place for a specific amount of time. Each agent also used a core autonomy pipeline, which included GPS-based state estimation, GPS-enabled local mapping, and a low-level controller capable of executing motor commands, for primitive action execution. In the following sections, we discuss each of the system components and associated challenges in more detail.

\subsection{Multiagent Collaborative Planning on Abstract Graphs}
\begin{figure}[t]
\centering
\includegraphics[trim={0 0cm 0 0},clip, width=\linewidth]{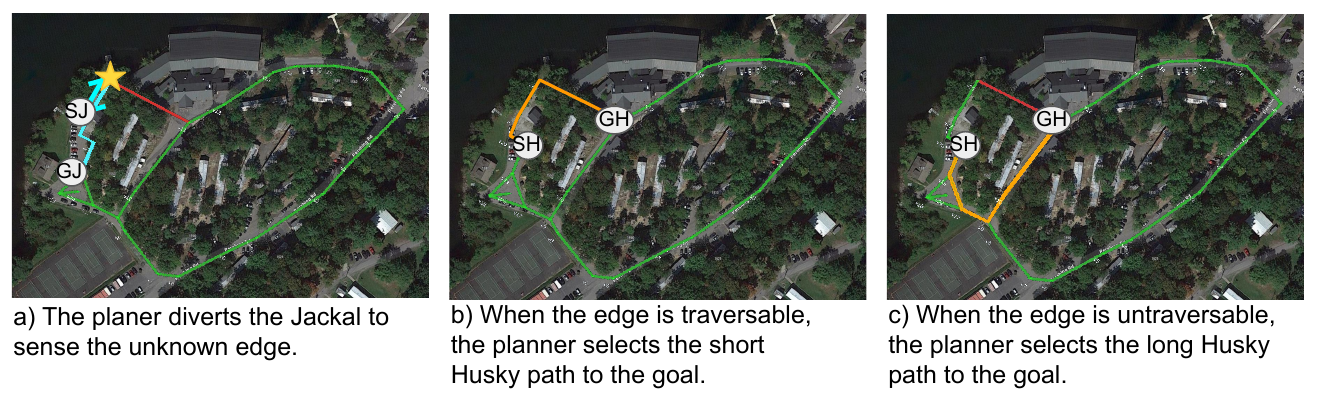}
\caption{An example collaborative team plan. The Jackal and Husky were tasked with navigating from SJ and SH to GJ and GH, respectively, by executing actions on a pre-defined abstract motion graph (green) where some edges had unknown traversabilities (red). In collaborative plans, agents took actions that were individually suboptimal to reduce the expected team makespan. Here, the Jackal diverted from its optimal path to the goal to sense and share the traversability of the unknown edge (a), while the Husky waited for the edge traversability information. Then, the Husky used the information to select the best traversable path to its goal, which reduced its expected plan cost (b-c).
}
\label{fig:collaborative_trajectory_example}
\end{figure}
The first level of the hierarchical planning system was collaborative team planning, based on an abstract motion graph of the environment and pre-specified agent starts and goals; an example collaborative plan is shown in Fig. \ref{fig:collaborative_trajectory_example}. Unfortunately, generating high-quality abstract graphs, which must efficiently (i.e., sparsely) represent high-quality, real-world agent traversals, but must also generate plans that can be grounded by real-world agents, is an active area of research \cite{veys2024generating}. In this work, we hand-generated an initial graph of the environment based on overhead imagery, where each node was associated with a GPS coordinate, and we fine-tuned the graph in the field to account for offsets and errors resulting from stale, low-fidelity overhead imagery. Additionally, we added deterministic edges between nodes where we expected that a valid, real-world traversal existed (e.g., when nodes were connected via roads), and we also added probabilistically traversable edges (with probability of being untraversable $\rho$) where we expected that a valid, real-world traversal may exist (e.g., when nodes were connected via a forested area). While edges and traversability probabilities were hand-defined in this work, we are interested in auto-generating graphs from remotely available information, like overhead imagery, in future work. 

We also used GPS throughout the deployment to correlate real-world agent locations with nodes in the graph. This was especially useful on startup, as we used GPS coordinates to define an initial alignment between team members in a global reference frame.  While using GPS was effective in our GPS-enabled environment, we are interested in exploring GPS-denied planning in the future. For example, it may be possible to generate a common team reference frame based on co-observed features on startup, or during planning using a distributed multiagent SLAM pipeline (e.g., \cite{kimera}).

\begin{figure}[t]
\centering
\includegraphics[width=.5\linewidth]{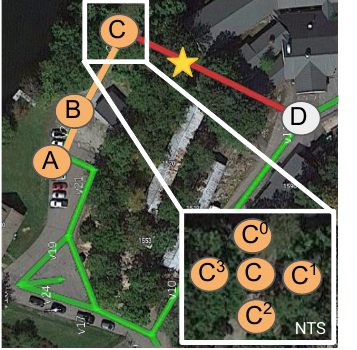}
\caption{An example macro-action (orange) that represents an agent first navigating from node A to B to C, and then sensing the traversability of the unknown edge C-D (gold star). The inlay depicts goal adjustment, which enables an agent to navigate to a location close to the original GPS location associated with the target node, if navigation to the target GPS location is deemed infeasible. This adjustment enables the agent to be robust to various types of uncertainty during planning, including minor pose drift, minor map noise, and small changes in the environment structure.}
\label{fig:macroaction}
\end{figure}

\subsection{Executing Macro-actions for Collaborative Planning}
\label{sect:macro-action-execution}
The second level of the hierarchical planner was the macro-action planner, which translated agent macro-actions into subgoals that could be executed in the real world. The planner processed macro-actions into sequences of executable primitive actions, monitored primitive action results, and combined the results of primitive action sequences into macro-action results (i.e., success or failure) that could be processed by the high-level collaborative planner. An example macro-action is shown in Fig. \ref{fig:macroaction}; the agent navigates from its current location, node A, to nodes B and C, and then to observe the traversability of the edge, C-D. 

\subsection{Primitive Actions}
Finally, the lowest level of the planner consisted of primitive action planners that generated executable robot commands. Specifically, the planners enabled robots to navigate along graph edges, sense the traversabilities of edges, and wait in place for a specific amount of time.
\subsubsection{Navigation Actions}
First, we developed a planner for navigating along a graph edge. Given the GPS coordinates corresponding to the source and target nodes of the edge, we set the agent navigation goal to be a region around the GPS coordinate of the target node; in the field trials, we used a circular region with a 3 meter radius. Then, we used a global planner, the EASL planner \cite{easl}, to generate a 2D plan between the agent's current location and the goal, where the planner used the agent's local occupancy map to avoid obstacles. Finally, we used an MPPI planner \cite{mppi} to generate agent motor commands to follow the EASL plan to the goal.

The primitive navigation planner was designed to be robust to various forms of real-world environmental uncertainty. By not constraining the agent to directly navigate along the graph edge (i.e., using a Euclidean path), we enabled the agent to react to local environment changes that did not impact global planning. For example, during field trials, an agent was able to navigate around an overgrown bush on the Euclidean path to its next node without impacting other levels of the planner. Additionally, the hierarchical approach enabled us to tune low-level planner parameters for individual agents without impacting the higher levels of the planning pipeline. This was especially useful for the MPPI planner, which used various parameters related to agent dynamics that varied across our vehicles.

\subsubsection{Observation Actions}
Second, we developed an observation action that enabled an agent to sense the local traversability of an adjacent edge. We used the current agent occupancy map and the EASL planner to generate a 2D plan from the current agent location to a point 20 meters along the unknown edge. Then, we the compared the EASL plan cost to the Euclidean distance between the two points. If the EASL plan cost exceeded 120\% of the Euclidean distance cost, the edge was marked untraversable; otherwise, the edge was marked traversable. Intuitively, this method tested if an agent could navigate along a graph edge without taking a significant detour, based on its local occupancy map. While the edge observation function used in our tests was primitive, other functions that report binary edge traversability based on local observations could be used in our pipeline. In future work, we are interested in generating binary observations based on other local traversability observation functions (e.g. \cite{guan2022tns}). We are also interested in exploring semantically informed observation functions (e.g., a function which marks an edge as traversable unless a suspicious barrel is present in the local map or agent camera images).

\subsubsection{Wait Actions}
Finally, we developed a wait action, which caused an agent to wait in place for a specified duration. This enabled an agent to wait in place to receive traversability observations from a teammate before making a decision about its next action. The wait action prevented costly backtracking when an agent did not have sufficient information to immediately make a high-quality planning decision, but a teammate was able to inexpensively sense the relevant information.

\subsection{Robust Macro-action Execution}
\label{sect:primitive_navigation_actions}
Unfortunately, planning abstractions are often imperfect, and discrepancies in planning abstractions at various levels of the planning hierarchy can lead to various types of planner failures.  For example, a low-level planning failure could be caused by the low-level planner, or it could be a result of a poor abstraction, like a poorly chosen GPS coordinate for a node (e.g., a coordinate that is in a bush). Hierarchical planners that are not robust to failures at various levels of of the planning system are likely to fail catastrophically.

\begin{figure}[t]
\centering
\includegraphics[width=0.8\linewidth]{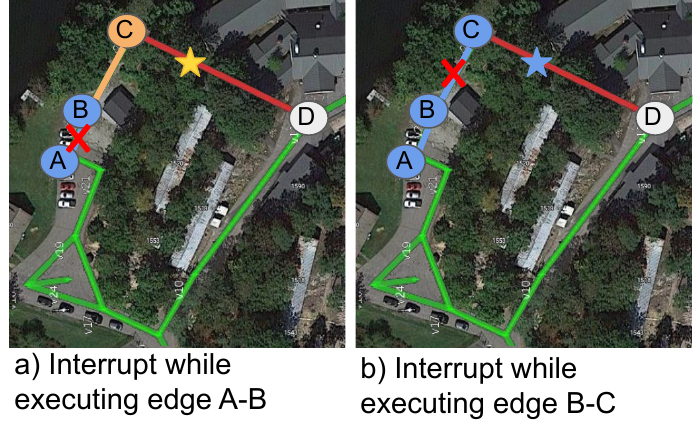}
\caption{Examples of macro-action interrupts (received when the agents are at the red Xs) and their impacts on planning. When an agent received an interrupt message during primitive action execution, it completed its current primitive navigation action and any directly succeeding sensing actions, then terminated the current macro-action, communicated with the base station, and waited to receive an updated plan.}
\label{fig:interrupt}
\end{figure}

In this work, after observing a number of primitive waypoint-based navigation failures, we updated the macro-action planner to be robust to such failures. Specifically, we modified the macro-action planner to re-attempt similar primitive navigation actions upon primitive action failure, rather than directly marking the macro-action as failed and reporting the failure to the abstract planner. The macro-action planner defined four alternative navigation goals for each node $(x, y)$: $(x + \delta, y), (x - \delta, y), (x, y + \delta)$, and $(x, y - \delta)$. If the agent failed to navigate to the original and all alternative goals, we assumed the primitive navigation action was unsuccessful, and the macro-action failed. However, if the agent successfully navigated to the original goal or any of the alternative goals, then the primitive  action was considered successful, and the macro-action continued\footnote{While this modification is similar to increasing the radius of the navigation goal region, we found that it resulted in more stable MPPI plans.}. In the experiments, we let $\delta = 0.5$m. While this macro-action planner modification was simple, it reduced the number of catastrophic failures experienced by the system, and demonstrated the importance of robustness across planning layers in hierarchical planning systems.

\subsection{Variable-Duration Macro-Actions}
\label{sect:interrupts}
Finally, we developed a method for coordinating teammates executing variable-duration macro-actions, like when agents execute macro-actions that represent traversals of graph edges with different weights. 
Unfortunately, it is not obvious how the team should plan when one agent successfully completes a macro-action, but another teammate is in the process of executing a macro-action.

We developed a planning approach that enabled the team to execute variable-length macro-actions online. Specifically, when any agent completed a macro-action, it sent the result of the macro-action (navigation success or failure, current state, and any traversability observations) to the centralized planner at the base station. Then, all other agents received an \textit{interrupt}, or a message indicating that the agent should terminate its current action as soon as possible, send action results to the base station, and wait to receive a new plan. If an agent received an interrupt message at the terminal state of a primitive action, then the macro-action terminated immediately, and the agent sent the macro-action result to the planner. Otherwise, the agent completed its current primitive navigation action and any observation actions that directly succeeded the navigation action, and then terminated the current macro-action and sent the action result to the centralized planner. For example, consider the agent in Fig. \ref{fig:interrupt} executing the macro-action of navigating from node A to node C, with the goal of sensing edge C-D. If the agent received an interrupt message when navigating between nodes A and B, the agent completed the primitive navigation action to node B, then terminated the macro-action, communicated with the base station, and waited for a replan. However, if the agent received an interrupt message when navigating between nodes B and C, the agent completed the primitive navigation action to node C, sensed the traversability of edge C-D, and then terminated the macro-action, communicated with the base station (including the traversability observation of edge C-D), and waited for a replan. Finally, once all agents terminated their current macro-actions, the centralized planner replanned for the entire team and sent new macro-actions to each agent. 

\begin{figure*}[tb] 
\centering
 \makebox[\textwidth]{\includegraphics[width=.8\paperwidth]{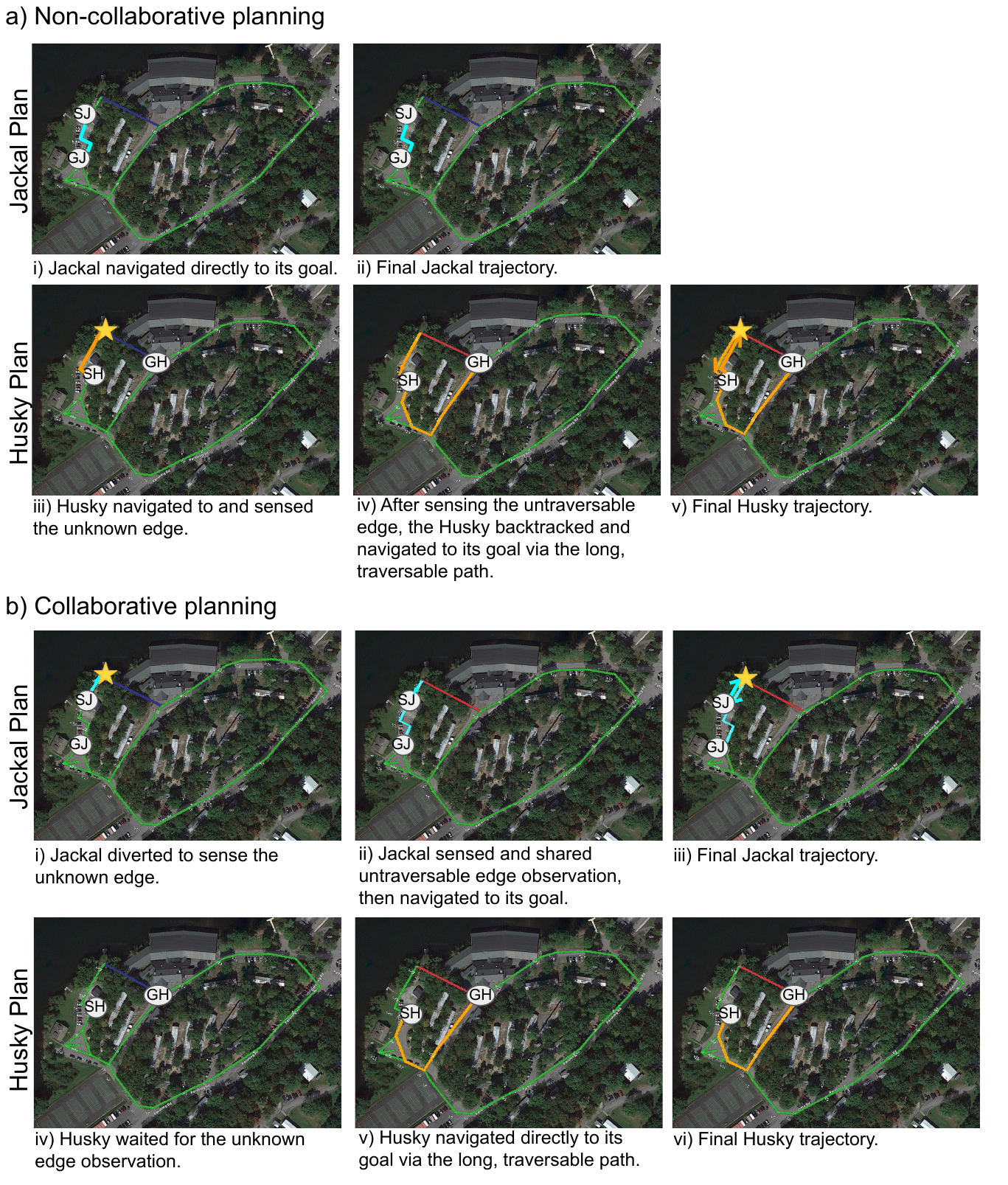}}
\caption{a) Graph-based plans selected by the non-collaborative baseline planner, visualized over the abstract navigation graph. The Jackal (blue) with start SJ and goal GJ navigated directly to its goal, while the Husky (orange) with start SH and goal GH navigated to the unknown edge, sensed that it was untraversable, and then backtracked to the goal via the long, traversable path. b) Graph-based plans selected by the collaborative planner, visualized over the abstract navigation graph. The Jackal (blue) with start SJ and goal GJ diverted to the gold star to sense the unknown edge, and shared the untraversable edge observation with the Husky. Then, the Jackal navigated to its goal. The Husky (orange) with start SH and goal GH waited to receive the edge observation from the Jackal, then navigated directly to its goal via the long, traversable path. Note that some trajectories were hand-annotated from raw agent outputs for clarity and due to agent/ground station communication drops.}
\label{fig:baseline_collab_plans}
\end{figure*}

The interrupt scheme enabled the team to quickly react to new information while ensuring that each agent remained in a valid configuration in the abstract graph. If an agent terminated its current action immediately after receiving an interrupt, its state may not be valid in the graph, and it would not be obvious how to generate a new graph-based plan for the agent. In future work, we are interested in developing methods that modify the abstract graph representation based on agent locations during interrupts. For example, if an agent is interrupted while navigating along a graph edge to a waypoint, it may be possible to add a node to the abstract graph at the agent interrupt location, and to split the current graph edge into two edges that represent the traversed and untraversed portions of the current edge, respectively. Finally, while this work does not directly aim to address the research problem of intermittent communications, the interrupt-based planner is robust to minor communication delays, and its lightweight messaging is tolerant of non-catastrophic outages.

\section{Results and Experiments}
We evaluated the system in real-world deployments in a structured outdoor environment; an overhead image of the environment is shown in Fig. \ref{fig:front_fig}-a. We evaluated our approach on a heterogeneous, two agent team consisting of a Clearpath Jackal and a Clearpath Husky.  Both agents were equipped with an Intel NUC11PH computer, a Microstrain 3DM-GX5-AHRS IMU, and a u-blox M8U GPS module, which provided GPS-based agent position estimates during the trials. Additionally, the Jackal was equipped with an Ouster OS1-32 LIDAR sensor, and the Husky was equipped with a Velodyne VLP-16 LIDAR sensor. Both agents were equipped with a core autonomy pipeline, which included GPS-based state estimation and an Omnimapper-based \cite{omnimapper}, GPS-enabled local mapping pipeline. All multiagent planning occurred on a centralized base station laptop, and the base station and agents communicated using a ROS multimaster system over a  Silvus radio network. Finally, it was assumed that the Jackal speed was 8x faster than the Husky speed.

\begin{figure}
\centering
\includegraphics[width=0.8\linewidth]{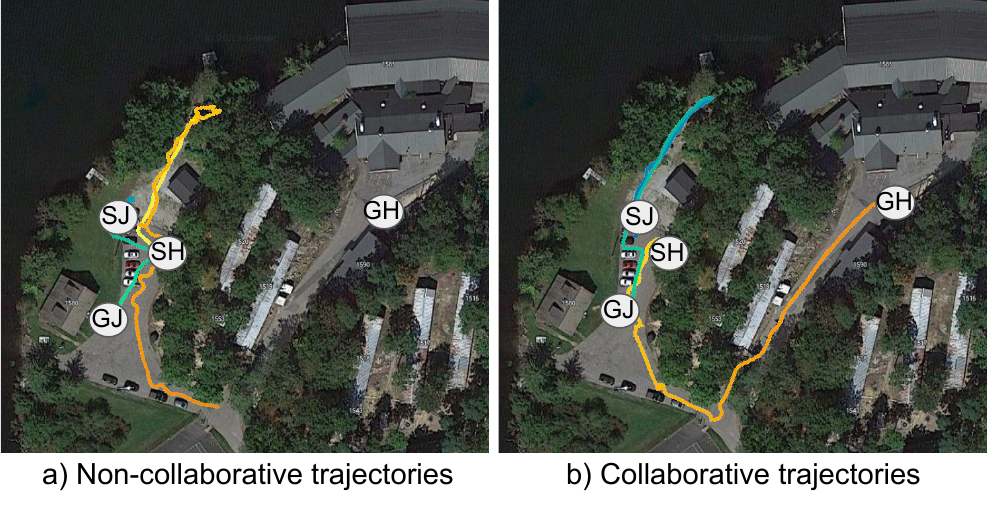}
\caption{Trajectories traversed by the Jackal (blue-green) and Husky (yellow-orange) while executing the graph-based plans described in Fig. \ref{fig:baseline_collab_plans}. Real-world trajectories are similar to the graph-based plans. Trajectories, starts, and goals are approximately scaled and hand-aligned on the overhead image.}
\label{fig:executed_trajectory}
\end{figure}

We evaluated the qualitative planning performance of the agents in various trials. In Fig. \ref{fig:baseline_collab_plans}, we compare the graph-based macro-actions selected by the collaborative and baseline planners during two successful planning trials. In the baseline, non-collaborative planning trial, the Jackal navigated directly to its goal, while the Husky attempted to navigate to its goal via the unknown edge (red). After the Husky arrived at the edge and sensed that it was untraversable, the Husky backtracked and navigated to its goal via a long, traversable path. In the collaborative planning trial, the Jackal diverted from the shortest path to the goal to sense the traversability of the unknown edge (red) at the gold star and relay the information to the Husky, while the Husky waited for information about the edge. After receiving the Jackal observation that the edge was not traversable, the Husky navigated directly to its goal via the long, traversable path. In Fig. \ref{fig:executed_trajectory}, we plot the trajectories traversed by the agents while executing the macro-action plans; trajectories, starts, and goals are approximately scaled and hand-aligned over the overhead image. Note that in the baseline trial, the Husky became stuck navigating up a steep hill while attempting to execute the final two edges to the goal. Also, in some trials, MPPI became stuck while trying to navigate around an obstacle (e.g. car, bush), and an operator briefly teleoperated the robot to enable the robot to continue to make progress towards completing the primitive action.  For this reason, we do not report the total execution time or distance traveled for the agents. However, it is possible to compare the plan costs for the agents in the abstract graph; we report the graph distance of each agent's planned trajectory, as well as the durations of planned wait actions, in Table \ref{tab:distance_wait_comparison}. In the collaborative trial, the Jackal travels further to sense the edge for the Husky, significantly reducing the graph distance traveled by the Husky. In the non-collaborative trials, the Jackal graph distance traveled is very low, but at the expense of a longer Husky trajectory. The results are consistent with the understanding that optimizing a team makespan requires a planner to trade off between the performances of the different agents.

\begin{table}
\centering
\small
\begin{tabular}{|p{1.7cm}|p{1.4cm}|p{1.4cm}|p{0.7cm}|p{0.7cm}|}
\hline
 & Graph Distance traveled, Husky (m) & Graph Distance traveled, Jackal (m)  & Husky Wait Time (s) & Jackal Wait Time (s) \\
\hline
Collaborative Planner & 156.71 & 100.04 & 10.0 & 0.0 \\
\hline
Non-Collaborative Planner & 257.97 & 25.83  & 0.0 & 0.0 \\
\hline
\end{tabular}
\caption{Agent wait times in seconds and plan costs in meters in the abstract graph when planning using the Collaborative and Non-Collaborative planners.}
\label{tab:distance_wait_comparison}
\end{table}

While we were able to demonstrate successful collaborative multiagent planning, we also observed various failures during testing. In Table \ref{tab:failure_types}, we summarize failure types which resulted in trial termination and their frequencies. 

\begin{table}
\centering
\small
\begin{tabular}{|p{1.0cm}|p{4.0cm}|}
\hline
Quantity & Type\\
\hline
4 & Low-level (EASL/MPPI) navigation failure \\
\hline
1 & Incorrect observation function output \\
\hline
\end{tabular}
\caption{Summary of failed trials.}
\label{tab:failure_types}
\end{table}

\section{Discussion}
While abstract planning representations are often necessary for efficiently solving complex planning problems, like problems in large or uncertain environments, or problems that involve teams of agents, significant challenges arise when executing abstract plans on real agents in real-world environments. First, building a representative abstraction of a real-world environment that is efficient for high-level planning, but that produces plans that can be translated into real-world executions, is an open research question. We are exploring methods for automatically generating abstract navigation graphs from polygonal environment models \cite{veys2024generating} and overhead images. While this will reduce operator burden, it is unlikely that the auto-generated graphs, like our hand-drawn graphs, will be fully representative of planning environments unless they have access to additional environment information, some of which may only be available during planning. We are exploring techniques to modify the navigation graph abstraction online based on local sensor data.

This work also demonstrated that robust hierarchical planning systems can overcome some of the challenges that arise from imperfect planning abstractions. Specifically, by increasing the robustness of planners at different levels of the planning hierarchy, we were able to overcome small discrepancies in the planning abstraction (like having a navigation waypoint in a bush next to a road, instead of on the road itself)  without causing catastrophic planning failures. In future work, we are interested in exploring other ways to make our hierarchical planning system more robust, including reconsidering traditional definitions of plan failure at each level in the planning hierarchy.

Finally, the proposed approach was developed for a centralized team with full communication. While our macro-action message passing scheme was lightweight, and our interrupt planner was capable of handling minor communication delays, in future work, we would like to modify the approach to explicitly handle more challenging communication environments.

\section{Conclusion}
As roboticists move towards deployments of bigger teams in larger, more complex environments, it will be necessary to develop abstract representations for complex planning problems that are still amenable to robust plan execution in real-world environments. In this work, we demonstrated the ability of  hierarchical planning systems to bridge the gap between abstract planning and real-world execution. We deployed a collaborative multiagent planner on a Jackal and Husky team using a hierarchical planning system, and demonstrated collaborative planning performance in a real-world structured outdoor environment. In future work, we are interested in improving methods for offline and online abstract planning representation generation, exploring methods for generating hierarchical planners that are robust at every planning level, and expanding our technique to more complex communication environments.

\addtolength{\textheight}{-12cm}   


\printbibliography

\end{document}